\documentclass[12pt]{article}
\usepackage{amssymb}
\usepackage[comma,numbers,square,sort&compress]{natbib}
\usepackage{amsmath}
\usepackage{natbib}
\usepackage{xcolor} 
\usepackage[hypertexnames=false,colorlinks=true,citecolor=blue,urlcolor=blue]{hyperref} 
\usepackage{graphicx}
\usepackage{float} 
\usepackage{amsfonts}
\usepackage{siunitx}
\usepackage{graphicx}
\usepackage{subcaption}
\usepackage{epstopdf}
\usepackage{setspace}
\usepackage[top=2.4cm,left=2.7cm,right=2.7cm,bottom=3cm]{geometry}

\title{\begin{center}
		\bfseries\singlespacing SOD-YOLO:  Enhancing YOLO-Based Detection of Small Objects in UAV Imagery
\end{center}}
\author{\parbox[c]{16cm}{\onehalfspacing \normalsize \centering ~\\[-0.4cm]
		Peijun WANG$^{1}$\qquad Jinhua ZHAO$^{1,2}\footnote{Correspondence: jzhao1101@ccnu.edu.cn}$\qquad \\ \footnotesize
		$^{1}$Faculty of Artificial Intelligence in Education, Central China Normal University, Wuhan 430079, China \\
		$^{2}$Faculty of Engineering \& Information Sciences, University of Wollongong, 2522 Wollongong, NSW, Australia\\
		[0.2cm]}
	\date{}
}

\begin{document}
\maketitle
\begin{abstract}
Small object detection remains a challenging problem in the field of object detection. To address this challenge, we propose an enhanced YOLOv8-based model, SOD-YOLO. This model integrates an ASF mechanism in the neck to enhance multi-scale feature fusion, adds a Small Object Detection Layer (named P2) to provide higher-resolution feature maps for better small object detection, and employs Soft-NMS to refine confidence scores and retain true positives. Experimental results demonstrate that SOD-YOLO significantly improves detection performance, achieving a 36.1\% increase in mAP$_{50:95}$ and 20.6\% increase in mAP$_{50}$ on the VisDrone2019-DET dataset compared to the baseline model. These enhancements make SOD-YOLO a practical and efficient solution for small object detection in UAV imagery. Our source code, hyper-parameters, and model weights are available at \href{https://github.com/iamwangxiaobai/SOD-YOLO}{https://github.com/iamwangxiaobai/SOD-YOLO}.
\end{abstract}
\begin{center}
\end{center}

\section{Introduction}
\label{introduction}

Unmanned aerial vehicles (UAVs), also known as drones, have revolutionized various fields, including wildlife monitoring, precision agriculture, search and rescue operations, and infrastructure inspection\cite{chen2022}. These applications often require detecting small objects, such as animals, pests, missing persons, or structural defects. The ability of UAVs to capture high-resolution imagery from otherwise inaccessible vantage points makes them invaluable assets. A pivotal task in UAV applications is object detection.

    Object detection in the UAV domain, especially small object detection, enhances UAV autonomy and effectiveness. It enables UAVs to identify and track small objects in real-time, contributing to various critical tasks. Among the myriad of object detection methodologies, two predominant strategies have emerged as the cornerstone of modern UAV-based systems: one-stage object detection and two-stage object detection.

    Two-stage object detection methods, exemplified by R-CNN (Region-based Convolutional Neural Network) and its variants \cite{cnn1,cnn4}, operate by first generating a set of candidate regions within the input image and then classifying each region individually. While effective, these approaches can be computationally intensive, posing challenges for real-time applications on resource-constrained UAV platforms. In contrast, one-stage object detection methods, such as YOLO (You Only Look Once) \cite{yolov1,yolov10}, adopt a more streamlined approach by directly predicting bounding boxes and class probabilities from the entire image in a single pass. This efficiency makes them particularly well-suited for UAV applications where real-time processing and low-latency responses are critical.
	
    YOLO's capability to detect multiple objects within a single frame enhances its utility in scenarios where comprehensive situational awareness is essential. However, YOLO does have inherent limitations. Its reduced accuracy compared to some Two-stage approaches, particularly in scenarios with small or densely packed objects, may be a concern. Additionally, YOLO's fixed grid cell structure may struggle to capture fine-grained details in objects, leading to suboptimal performance in certain contexts. Furthermore, YOLO's reliance on predefined anchor boxes may limit its adaptability to diverse environments and object scales commonly encountered in UAV operations.
	
    In this paper, we propose an enhanced YOLOv8 algorithm, termed SOD-YOLO (Small Object Detection YOLO), aimed at improving small object detection in UAV imagery. Our modifications address the challenges posed by the diminutive size and complex backgrounds of small objects. The experiment result shows that our SOD-YOLO has improved the performance of YOLOv8 on VisDrone2019-DET \cite{visdrone2019} dataset by \SI{36.1}{\percent} (mAP\textsubscript{$50:95$}) and \SI{20.6}{\percent} (mAP\textsubscript{$50$}). 
    Our main contributions can be summarized as follows: 
    \begin{itemize}
        \item The integration of ASF mechanism \cite{asf}, a novel attentional fusion strategy in the neck, enhances the model's capability to handle variations in size and complex backgrounds.
        \item The introduction of the P2 layer involves a dedicated small object detection head that utilizes high-resolution feature maps to significantly improve accuracy in detecting small objects.
        \item  Implementation of Soft-NMS \cite{softnms}, an advanced technique for refining confidence scores, effectively retains true positives and enhances the overall performance of object detection.
    \end{itemize}

\section{ Related Work}

	\subsection{Real-time Object Detectors}
	YOLO stands out among object detection methods for its pioneering one-stage, real-time capable neural network approach. Unlike traditional two-stage detectors like R-CNN, YOLO directly predicts bounding boxes and class probabilities from the entire image in a single pass. This design significantly reduces computation time, making it well-suited for real-time applications. YOLO has undergone several iterations, with YOLOv5 and YOLOv8 being notable advancements, improving both speed and accuracy. YOLO's efficiency has made it a popular choice in various domain, where low latency and high throughput are critical.
	
	\subsection{Previous Works on Small Object Detection}

            \subsubsection{Network Architecture Enhancements}
     In the field of object detection, significant advancements have been made through improvements in network architecture. Since the advent of YOLOv1, the YOLO series has consistently led real-time object detection, driven by continuous enhancements in both backbone networks and detection layers. Traditional backbones like Darknet have been widely utilized, balancing computational efficiency with detection performance. More recent developments include advanced backbones such as CSPNet \cite{cspnet} and EfficientNet \cite{efficientnet}, which offer improved feature extraction and network efficiency. Despite these advancements, challenges remain in accurately detecting small objects, as conventional backbones and detection layers often fail to capture the fine-grained features necessary for high accuracy. Additionally, small objects can be easily overlooked or confused with background noise, leading to lower detection rates and higher false negatives.
     
     Building on these foundations, our work introduces several key architectural improvements within the YOLOv8 framework. We propose the ASF network as a novel neck, designed to dynamically adjust spatial filters based on object scale and context. This adaptation enhances feature extraction, capturing detailed spatial information crucial for small object detection. Furthermore, we integrate a dedicated small object detection layer, specifically tailored to focus on fine-grained features that conventional detection layers might overlook. These combined enhancements significantly boost the model's sensitivity and accuracy in detecting small objects,  resulting in superior performance in challenging environments.
            \subsubsection{Post-Processing Techniques}
        
        Post-processing techniques are essential for refining the results generated by object detection models. Traditional Non-Maximum Suppression (NMS) techniques, though effective, often discard true positive detections due to their hard suppression criteria, which is particularly problematic for small object detection where overlapping bounding boxes can result in significant information loss. This issue is exacerbated in densely packed scenes where small objects are close together. Moreover, the hard thresholding approach of NMS can lead to suboptimal precision-recall balance, making it difficult to achieve high accuracy across diverse object sizes and densities. Recent advancements, such as Soft-NMS, have been introduced to address these issues. Soft-NMS gradually reduces the confidence scores of overlapping bounding boxes instead of eliminating them outright, thereby maintaining high recall rates and improving precision. This approach has shown promise in enhancing the overall performance of object detection models, particularly in dense and cluttered environments.
                
        In our approach, we integrate Soft-NMS into the YOLOv8 framework to enhance small object detection performance. Soft-NMS allows for a more nuanced suppression of overlapping detections, reducing the risk of discarding true positives. By incorporating this technique, our model achieves better performance in small object detection, ensuring more accurate and reliable results, especially in scenarios with densely packed objects. This enhancement significantly contributes to the robustness and effectiveness of our detection system, ensuring high precision and recall in real-world applications.
\section{Approach}\label{sec4}
\subsection{Integration of ASF Mechanism}

\begin{figure*}[htbp]
    \centering
    \includegraphics[width=\textwidth]{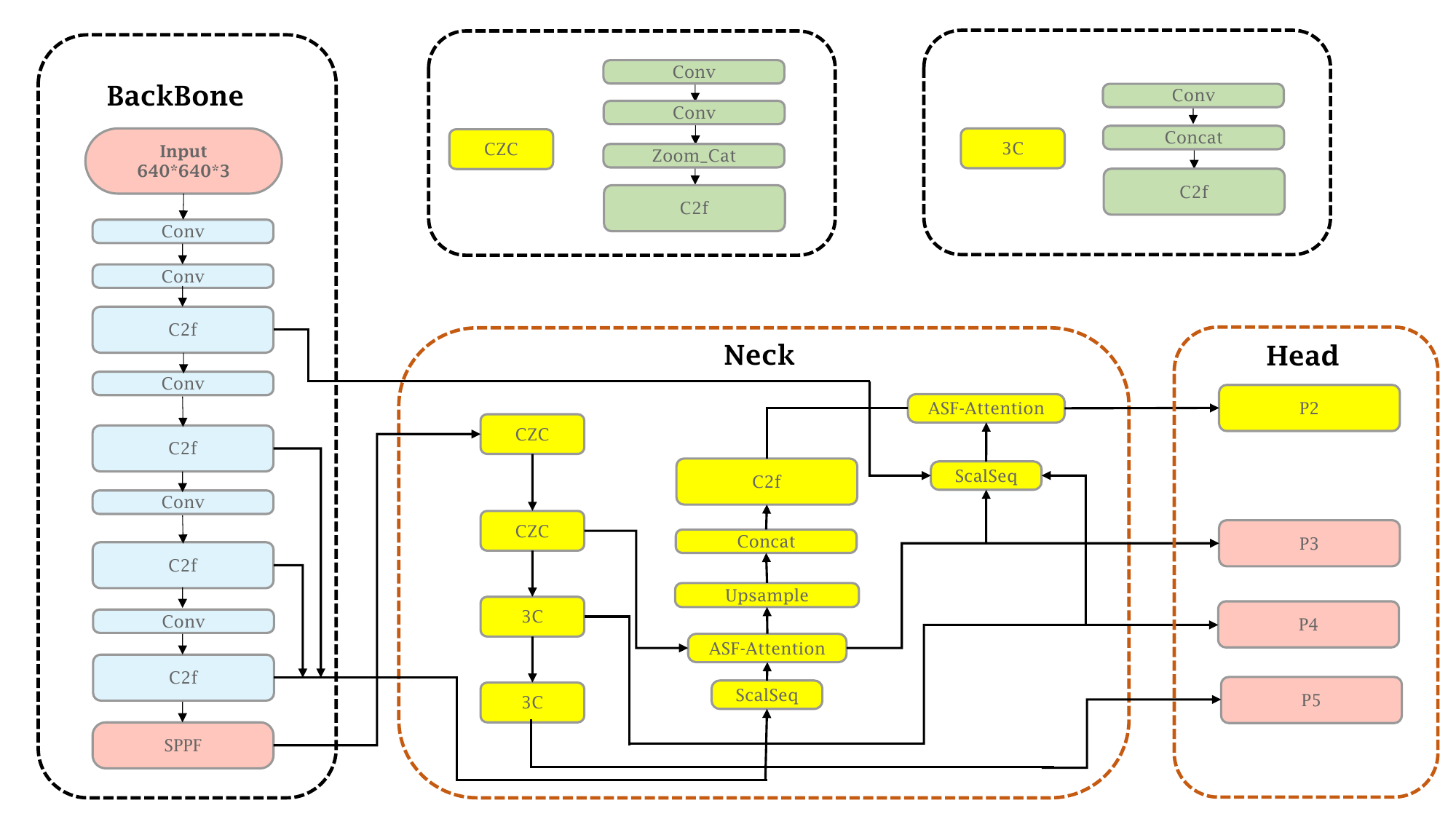} 
    \caption{Network Structure: SOD-YOLO}
    \label{SOD-YOLO}
\end{figure*}

The standard YOLOv8 architecture adopts a sequential neck design that progressively downsamples and enriches feature maps through Convolution (Conv), C2f (an enhanced residual block), and SPPF (Spatial Pyramid Pooling Fusion) layers\cite{sppf}. Its head concatenates multi-scale features (P3, P4, P5) and processes them through a detection layer for object prediction.

In our work on small object detection from UAV perspectives, we innovatively integrate the Adaptive Scale Fusion (ASF) mechanism—originally developed for cell instance segmentation~\cite{asf}—into the YOLOv8 framework, as shown in Figure~\ref{SOD-YOLO}. Our improvements are primarily focused on a novel Scale Sequence (ScalSeq) Feature Fusion module and an ASF Attention-Based Refinement block.

\textbf{Scale Sequence (ScalSeq) Feature Fusion:}  
In contrast to the simple concatenation strategy employed in YOLOv8, our neck incorporates two instances of the proposed Scale Sequence (ScalSeq) module at critical stages of the feature aggregation process. Each ScalSeq unit receives three input feature maps from different scales—specifically, P3, P4, and P5 as defined in the YOLOv8 architecture. Initially, each feature map is passed through a $1 \times 1$ convolution to unify their channel dimensions. Subsequently, the P4 and P5 feature maps are upsampled via nearest-neighbor interpolation to match the spatial resolution of P3.

Once spatially aligned, the three feature maps are treated as a sequence along a newly introduced ``scale'' dimension. A 3D convolution with a kernel size of $(1, 1, 1)$ is then applied along this dimension, enabling the model to learn cross-scale semantic fusion and contextual representation. This operation is followed by batch normalization along the scale dimension and a LeakyReLU activation function. Finally, a 3D max pooling with a kernel size of $(3, 1, 1)$ is employed to compress the scale dimension, resulting in a standard 2D feature map. The output of the ScalSeq module replaces the conventional concatenation mechanism typically used in the YOLOv8 neck, providing a more expressive and structured fusion of multi-scale information.

\textbf{ASF Attention-Based Refinement:}
Following each ScalSeq module, we introduce a lightweight attention block named attention\_model to further refine the fused features. This block implements the Adaptive Scale Fusion (ASF) attention mechanism. It takes two feature maps as input and sequentially applies a channel attention module to the first input feature map. The result of this channel attention is then added to the second input feature map. Finally, a local attention module is applied to the summed feature map. This sequential application of channel and local attention allows the network to selectively emphasize informative features and suppress noise prevalent in UAV imagery. The resulting ASF-enhanced features are then used in subsequent layers of the neck or as part of skip connections, improving the quality of upsampled features before the final prediction stage.

These modifications to the neck structure enable a more sophisticated fusion of multi-scale features compared to YOLOv8, allowing SOD-YOLO to better preserve the detailed information necessary for detecting small objects while maintaining computational efficiency suitable for real-time UAV applications.

    \subsection{Small Object Detection Layer}

    To enhance the detection of small objects in UAV imagery, we introduce an additional detection head at the P2 level of the network, as illustrated in Figure \ref{SOD-YOLO}. Unlike the standard detection heads at P3, P4, and P5, which correspond to deeper and more abstract feature representations with lower spatial resolution, the P2 layer operates on higher-resolution feature maps derived from earlier stages of the backbone. These shallow layers retain more fine-grained spatial information, such as object boundaries, textures, and edges, which are critical cues for detecting small-scale targets in aerial scenes\cite{ssd}\cite{zhao2019review}.
    
    The rationale behind incorporating a P2 detection layer lies in the inherent characteristics of small objects: they often occupy very few pixels in high-altitude UAV imagery and are susceptible to being overwhelmed or diluted by downsampling in deeper layers\cite{gan}. While P3 already offers moderate resolution, it may still lose essential local details due to strided convolutions. In contrast, P2 preserves finer spatial granularity and captures low-level visual features that are better suited for identifying small targets, especially under complex backgrounds and occlusions.
    
    To effectively utilize these high-resolution features, we first upsample selected feature maps from the backbone and concatenate them with shallow-layer outputs to form a rich feature set that combines both spatial detail and contextual depth. This composite feature is refined through a C2f module—an enhanced residual block designed to preserve critical edge and shape information while enabling effective feature fusion\cite{cspnet}.
    
    Subsequently, the processed P2 feature maps are passed through a Scale Sequence (ScalSeq) module, which aggregates multi-scale information and emphasizes small object representations. Finally, the enriched P2 output is integrated alongside existing P3–P5 detection heads to form a multi-resolution detection head ensemble. This design ensures comprehensive coverage of object scales, particularly improving the model’s sensitivity and precision for detecting small and densely distributed objects.
    
    Overall, the addition of the P2 detection layer equips the model with early-stage spatial cues, enhancing its robustness in small object detection tasks typical in UAV-based visual applications. It bridges the gap between high-resolution local features and deep semantic understanding, leading to significantly improved detection accuracy in cluttered and scale-variant aerial environments.

\begin{table*}[h]
  \centering
  \caption{Comparison of different object detectors on VisDrone2019-DET-val. }
  \label{tab1}
  \begin{tabular}{c c c c c}
    \hline
    Model & mAP$_{50:95}$ & mAP$_{50}$ & Param (M) & FLOPs (G) \\ \hline
    YOLOv5-m(2020) \cite{yolov5} & 0.225 & 0.385 & \textbf{20.8} & \textbf{48.1} \\
    YOLOv7-m(2022) \cite{yolov7} & 0.265 & 0.472 & 36.9 & 103.5 \\
    YOLOv8-m(2023) & 0.258 & 0.436 & 25.8 & 78.7 \\
    YOLOv9-gelan-c(2024) \cite{yolov9} & 0.305 & 0.489 & 25.2 & 101.8 \\
    YOLOv10-l(2024) \cite{yolov10} & 0.286 & 0.462 & 25.7 & 126.4 \\
    SOD-YOLO (Ours) & \textbf{0.351} & \textbf{0.526} & 22.6 & 94.9 \\
    \hline
  \end{tabular}
\end{table*}

\begin{table*}[h]
  \centering
  \caption{Ablation Study}
  \label{tab3}
  \begin{tabular}{l c c c}
    \hline
    Model & mAP$_{50:95}$ & mAP$_{50}$ & FLOPs (G) \\ \hline
    Baseline & 0.258 & 0.436 & 78.7 \\
    +ASF & 0.265 \textcolor{red}{(+0.007)} & 0.440 \textcolor{red}{(+0.004)} & 82.7 \textcolor{green}{(+4.0)} \\
    +ASF+P2 & 0.294 \textcolor{red}{(+0.036)} & 0.476 \textcolor{red}{(+0.040)} & 94.9 \textcolor{green}{(+16.2)} \\
    +ASF+P2+SoftNMS & 0.352 \textcolor{red}{(+0.094)} & 0.526 \textcolor{red}{(+0.090)} & 94.9 \\
    \hline
  \end{tabular}
\end{table*}

\subsection{Soft-NMS Integration}
NMS is a key post-processing technique in object detection models, designed to remove redundant overlapping bounding boxes. The traditional NMS algorithm reduces the score of a detection frame to 0 when the IoU between the current detection frame and the frame with the highest score exceeds a certain threshold\cite{tranms}. This method can result in missing object frames with significant overlap, leading to a reduced recall rate for small objects.

To address the issue of potentially discarding detection frames containing objects, our study employs the soft-NMS\cite{softnms} algorithm. Unlike traditional NMS, soft-NMS adjusts the score of the current detection frame by applying a weighting function, which diminishes the scores of neighboring frames that overlap with the highest scoring frame. The greater the overlap with the highest scoring frame, the faster the score decays. This approach prevents the removal of frames containing objects and avoids situations where two similar frames both detect the same object. The soft-NMS algorithm used in our study accounts for the overlap between detection frames, reducing the likelihood of false negatives compared to the traditional NMS algorithm, and thus improving the accuracy and reliability of small object detection.

The processing of the conventional NMS algorithm can be visually expressed by the fractional reset function in Equation \eqref{eq:nms}.

\begin{equation}
S_i =
\begin{cases} 
s_i & \text{IoU}(A, B_i) < N_t \\
0 & \text{IoU}(A, B_i) \geq N_t 
\end{cases}
\label{eq:nms}
\end{equation}

The soft-NMS algorithm can be visually expressed as a fractional reset function as shown in Equation \eqref{eq:soft-nms}.

\begin{equation}
S_i =
\begin{cases} 
s_i & \text{IoU}(A, B_i) < N_t \\
s_i(1 - \text{IoU}(A, B_i)) & \text{IoU}(A, B_i) \geq N_t 
\end{cases}
\label{eq:soft-nms}
\end{equation}

In Equations \eqref{eq:nms} and \eqref{eq:soft-nms}, \( S_i \) denotes the score of the \(i\)-th detection frame; \( A \) indicates the detection frame with the highest confidence in the region of interest; \( B_i \) represents the \(i\)-th detection box; IoU denotes the overlap of the \(i\)-th detection frame with \( A \); and \( N_t \) indicates the calibrated overlap threshold.

\section{Experiments}
\subsection{Experiment Setup Overview}
To validate the effectiveness and robustness of our proposed SOD-YOLO model for small object detection in UAV imagery, we conduct extensive experiments under various scenarios. Our experimental setup is designed with the following objectives:

\begin{itemize}
    \item \textbf{Evaluation Goals:} Assess the performance of SOD-YOLO in detecting small and densely packed objects in complex aerial scenes, and verify its improvements over the baseline YOLOv8-m model.
    \item \textbf{Evaluation Metrics:} We primarily report mean Average Precision (mAP) at IoU thresholds of 0.5 (mAP@0.5) and 0.5:0.95 (mAP@0.5:0.95), along with metrics for small object detection as defined by the respective datasets.
    \item \textbf{Comparison Baselines:} We compare SOD-YOLO against the baseline YOLOv8-m, as well as other state-of-the-art methods when applicable. To ensure fairness, all models are trained and evaluated under the same settings.
\end{itemize}

\subsection{Datasets}

\textbf{VisDrone2019-DET.} We evaluate our model on the widely used VisDrone2019-DET dataset\cite{visdrone2019}, a comprehensive benchmark for UAV-based object detection. The dataset contains 10,209 static images and 261,908 video frames from 288 drone video clips, captured across various cities in China. It includes diverse scenes (urban and rural), object scales, and densities. The static image set is divided into 6,471 training, 548 validation, and 1,610 testing images.

VisDrone defines 10 object categories: \textit{pedestrian, people, bicycle, car, van, truck, tricycle, awning-tricycle, bus}, and \textit{motor}. Notably, over 75\% of the annotated objects occupy less than 0.1\% of the image area, highlighting the dominance of small and tiny objects. Object distribution also shows a strong central bias, justifying the use of center-focused augmentations.

\subsection{Implementation Details}

\textbf{Training Environment.} All experiments are conducted on a workstation equipped with a single NVIDIA RTX 4090 GPU. The model backbone is CSPDarknet53. Training is performed using the Stochastic Gradient Descent (SGD) optimizer with a cosine learning rate schedule and warm-up strategy.

\textbf{Main Hyperparameters.}
\begin{itemize}
  \item Initial learning rate: linearly warmed up to 0.005 in 3 epochs
  \item Weight decay: 0.0005
  \item Momentum: 0.937
  \item Total training epochs: 200
  \item Input image size: $640 \times 640$
  \item Batch size: 8
\end{itemize}

For more hyperparameter settings and implementation details, please refer to our GitHub repository: \href{https://github.com/iamwangxiaobai/SOD-YOLO}{https://github.com/iamwangxiaobai/SOD-YOLO}. 

\begin{figure*}[htbp]
  \centering
  \begin{subfigure}{0.45\textwidth}
    \includegraphics[width=\linewidth]{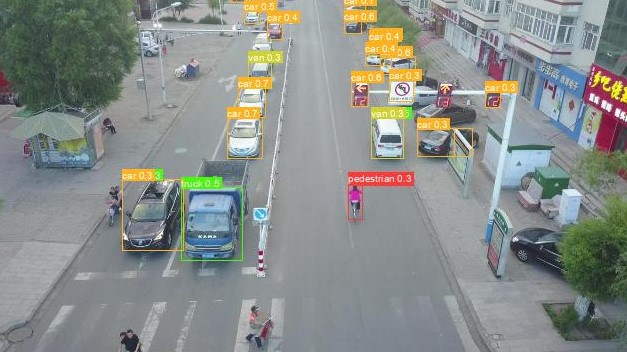}
    \caption{YOLOv8}
    \label{fig:baseline}
  \end{subfigure}
  \hfill
  \begin{subfigure}{0.45\textwidth}
    \includegraphics[width=\linewidth]{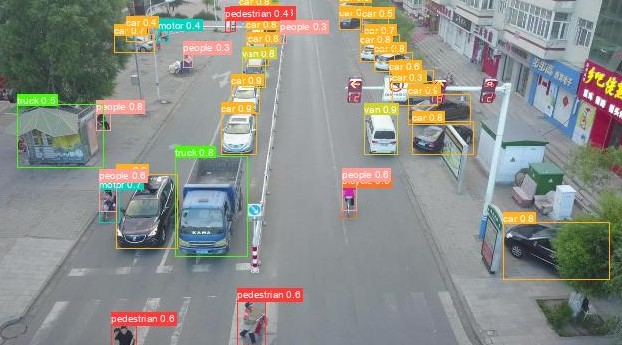}
    \caption{SOD-YOLO (Ours)}
    \label{fig:improved}
  \end{subfigure}
  \caption{Comparison of detection performance between YOLOv8 and SOD-YOLO.}
  \label{fig:comparison}
\end{figure*}

\subsection{Evaluation Criterion}
The common criteria used to evaluate the performance of an object detection algorithm include Intersection over Union (IoU), Precision, Recall, and Mean Average Precision (mAP). The detailed definitions are listed below:

\begin{enumerate}
    \item \textbf{IoU (Intersection over Union)}: IoU is calculated by taking the overlap area between the predicted region (A) and the actual ground truth (B) and dividing it by the combined area of the two. The formula can be expressed as:
    \begin{equation}
    \text{IoU} = \frac{|A \cap B|}{|A \cup B|}
    \end{equation}
    The value of IoU ranges from 0 to 1. A larger IoU indicates a more precise prediction. A lower numerator (intersection) value suggests that the prediction failed to accurately overlap the ground truth, while a larger denominator (union) implies a broader predicted region, resulting in a lower IoU.
    
    \item \textbf{Precision}: Precision represents the proportion of correctly predicted positive samples out of all predicted positive samples. It is defined as:
    \begin{equation}
    \text{Precision} = \frac{\text{True Positives}}{\text{True Positives} + \text{False Positives}}
    \end{equation}
    
    \item \textbf{Recall}: Recall represents the proportion of actual positive samples that were correctly predicted. It is calculated as:
    \begin{equation}
    \text{Recall} = \frac{\text{True Positives}}{\text{True Positives} + \text{False Negatives}}
    \end{equation}
    
    \item \textbf{mAP (Mean Average Precision)}: Average Precision (AP) is a measure of the Precision scores at different recall thresholds along the Precision-Recall (PR) curve. mAP is the mean AP across all object classes. Specifically, mAP$_{50}$ refers to the mAP when IoU is 0.5, while mAP$_{50:95}$ represents the mean mAP for IoU thresholds ranging from 0.5 to 0.95.
\end{enumerate}

\subsection{Results and Comparison}
	We choose YOLOv8-m as our baseline model and compare our detector with several state-of-the-art object detectors on the VisDrone2019-DET dataset, which primarily consists of small objects.  The results on VisDrone2019-DET-val are shown in Table \ref{tab1}.  After comparing the experimental outcomes between YOLOv8-m and the improved SOD-YOLO algorithm, it is evident that our algorithm significantly outperforms YOLOv8-m, particularly in detecting small targets. SOD-YOLO achieves an mAP$_{50:95}$ of 0.351, which represents a 0.093 improvement over YOLOv8-m’s 0.258. Similarly, the mAP$_{50}$ of SOD-YOLO reaches 0.526, a 0.09 increase from YOLOv8-m’s 0.436. This demonstrates that SOD-YOLO enhances the detection of small objects, especially at higher IoU thresholds, indicating better precision and recall compared to the baseline.Additionally, when compared to other models such as YOLOv9-gelan-c and YOLOv10-l, SOD-YOLO remains competitive. It surpasses YOLOv9-gelan-c in both mAP$_{50:95}$ (0.351 vs. 0.305) and mAP$_{50}$ (0.526 vs. 0.489), despite YOLOv9 having similar parameter counts. This highlights the efficiency of our model in balancing detection accuracy and computational cost. Furthermore, SOD-YOLO maintains a relatively low parameter count (22.6M) and competitive FLOPs (94.9G), providing a more efficient solution than models like YOLOv7-m (36.9M parameters, 103.5G FLOPs) and Edge-YOLO. Thus, SOD-YOLO offers improved performance for small object detection with modest computational requirements.

    Figure \ref{fig:comparison} presents the detection results of the baseline YOLOv8-m model and the proposed SOD-YOLO model on the same urban UAV image. Compared to the baseline, the SOD-YOLO model demonstrates a significantly improved capability in detecting small and partially occluded objects. Notably, SOD-YOLO successfully identifies multiple pedestrians on the left side of the road that are either missed or misclassified by YOLOv8-m. For example, pedestrians near the crosswalk and a small child close to the curb are correctly detected by SOD-YOLO, while YOLOv8-m fails to detect them or mislabels them as other classes.
    
    Additionally, SOD-YOLO detects more distant vehicles and motors with higher confidence, particularly in the middle to far end of the street, where YOLOv8-m either omits them or assigns low-confidence bounding boxes. The enhanced bounding box localization and class precision of SOD-YOLO are evident in its correct distinction between trucks, vans, and cars, whereas YOLOv8-m shows occasional confusion between similar-sized vehicle classes.
    
    These visual comparisons affirm that the proposed enhancements in SOD-YOLO—such as the introduction of the ASF module, P2 detection layer, and Soft-NMS—contribute to superior performance in challenging UAV scenes involving dense, small, and occluded object detection.

\subsection{Ablation Study}
We conducted an ablation study to evaluate the contributions of each component in our proposed SOD-YOLO model, with the results presented in Table~\ref{tab3}. In the table, values highlighted in red indicate performance improvements, while those in green represent declines. The baseline model, YOLOv8-m, achieved an mAP\textsubscript{$50:95$} of 0.258 and an mAP\textsubscript{$50$} of 0.436 with a computational complexity of 78.7 GFLOPs. By introducing the Attentional Fusion Strategy (ASF) in the neck of the model, we observed improvements in mAP\textsubscript{$50:95$} to 0.265 (\textbf{+0.007}) and mAP\textsubscript{$50$} to 0.440 (\textbf{+0.004}), though this modification increased FLOPs to 82.7G. The subsequent addition of the P2 layer, which provides a dedicated detection head for small objects using high-resolution feature maps, led to a further increase in mAP\textsubscript{$50:95$} to 0.294 (\textbf{+0.036}) and mAP\textsubscript{$50$} to 0.476 (\textbf{+0.040}), albeit with a rise in FLOPs to 94.9G. Finally, the integration of Soft-NMS, which refines the confidence scores to better retain true positives, significantly boosted the performance, achieving an mAP\textsubscript{$50:95$} of 0.352 (\textbf{+0.094}) and an mAP\textsubscript{$50$} of 0.526 (\textbf{+0.090}), with no further increase in computational complexity, maintaining FLOPs at 94.9G. These results demonstrate that each component contributes to the overall improvement of the model, with the combined effect of ASF, P2, and Soft-NMS delivering a substantial performance boost.

\section{Conclusions}

In this paper, we presented SOD-YOLO, an enhanced YOLOv8 algorithm tailored for small object detection in UAV imagery. Our modifications, including the integration of the ASF mechanism, the introduction of a dedicated P2 detection layer, and the implementation of Soft-NMS, significantly improved detection performance. The ASF mechanism enhances spatial feature extraction by focusing on key regions, allowing the model to better detect small and densely packed objects. The addition of the P2 detection layer captures finer details at a lower feature scale, improving small object detection, while Soft-NMS reduces false positives by softly penalizing overlapping boxes rather than discarding them. Experiments on the VisDrone2019-DET dataset demonstrated a 36.1\% increase in mAP$_{50:95}$ and a 20.6\% increase in mAP$_{50}$ over the baseline YOLOv8, showcasing the effectiveness of our approach in handling small objects in complex UAV scenes.

Despite the significant improvements achieved, there are still some limitations in our work. First, although the ASF mechanism and the P2 detection layer effectively enhance small object detection accuracy, the increased computational complexity of the model impacts its real-time performance. This limitation may restrict the deployment of the model, especially in scenarios with limited computational resources. Therefore, future work could focus on optimizing the model's computational efficiency through lightweight design to ensure performance in real-time applications.

\section*{Acknowledgment}
This work is supported by National Natural Science Foundation of China, Grant No. 72101189.

\end{document}